\documentclass{article}
\usepackage{ijcai19}

\usepackage{times}
\usepackage{soul}
\usepackage{url}
\usepackage[hidelinks]{hyperref}
\usepackage[utf8]{inputenc}
\usepackage[small]{caption}
\usepackage{graphicx}
\usepackage{amsmath}
\usepackage{booktabs}
\usepackage{algorithm}
\usepackage{algorithmic}
\usepackage{multirow}
\usepackage{subcaption}
\usepackage{booktabs}
\title{Multi-Task Learning with Contextualized Word Representations \\for Extented Named Entity Recognition}

\author{
Thai-Hoang Pham$^{1,2}$
\and
Khai Mai$^2$\and
Nguyen Minh Trung$^2$\and
Nguyen Tuan Duc$^2$\and\\
Danushka Bolegala$^3$\and
Ryohei Sasano$^4$\And
Satoshi Sekine$^5$
\affiliations
$^1$Ohio State University\\
$^2$Alt Inc\\
$^3$University of Liverpool\\
$^4$Nagoya University\\
$^5$Riken AIP
\emails
pham.375@osu.edu,
\{mai.tien.khai, nguyen.minh.trung, nguyen.tuan.duc\}@alt.ai,
danushka.bollegala@liverpool.ac.uk,
sasano@i.nagoya-u.ac.jp,
satoshi.sekine@riken.jp
}

\begin{document}

\maketitle

\begin{abstract}
Fine-Grained Named Entity Recognition (FG-NER) is critical for many NLP applications. While classical named entity recognition (NER) has attracted a substantial amount of research, FG-NER is still an open research domain. The current state-of-the-art (SOTA) model for FG-NER relies heavily on manual efforts for building a dictionary and designing hand-crafted features. The end-to-end framework which achieved the SOTA result for NER did not get the competitive result compared to SOTA model for FG-NER. In this paper, we investigate how effective multi-task learning approaches are in an end-to-end framework for FG-NER at different aspects. Our experiments show that using multi-task learning approaches with contextualized word representations can help an end-to-end neural network model achieve SOTA results without using any additional manual effort for creating data and designing features. 
\end{abstract}

\section{Introduction}
\label{sec:introduction}
\begin{figure}[t]
\centering
\begin{subfigure}{0.45\textwidth}
\includegraphics[width=1\linewidth]{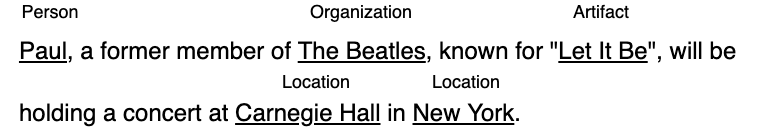}
\subcaption{NER result}
\label{fig:0a}
\end{subfigure}\vspace{2mm}
\begin{subfigure}{0.45\textwidth}
\includegraphics[width=1\linewidth]{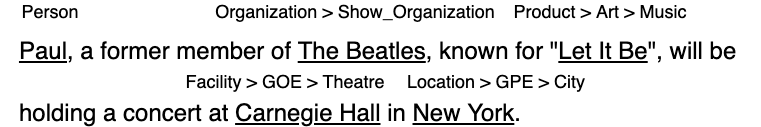}
\subcaption{FG-NER result}
\label{fig:0b}
\end{subfigure}
\caption{Example of NER and FG-NER.}
\label{fig:0}
\end{figure}

\paragraph{}
Fine-grained named entity recognition (FG-NER) is a special kind of named entity recognition (NER) that focuses on identifying and classifying a large number of entity categories. In traditional NER task, often less than eleven named entity (NE) categories are defined. For example, in two shared
tasks, CoNLL 2002 and CoNLL 2003~\cite{Tjong:2002,Tjong:2003}, there were only four NE types considered: Person, Location, Organization, and Miscellaneous. From these shared tasks, ten NE categories were defined for Twitter texts~\cite{Ritter:2011}. The FG-NER, on the other hand, handles hundreds NE categories which are the fine-grained classification of coarse-grained categories. In particular,~\cite{Sekine:2002,Sekine:2008} proposed the entity hierarchy which contains 200 NE categories designed manually. Meanwhile,~\cite{Ling:2012,Yosef:2012,Gillick:2014} used unsupervised methods for creating FG-NER category from knowledge bases such as Freebase~\cite{Bollacker:2008} and YAGO~\cite{Suchanek:2007}. Figure~\ref{fig:0} shows an example when identifying and classifying NE by traditional NER and FG-NER systems.

\paragraph{}
While there have been many methods proposed for classical NER~\cite{Zhou:2002,Mccallum:2003,Ma:2016,Pham:2017}, FG-NER is still an open research domain. Unlike NER, the current SOTA model for FG-NER~\cite{Mai:2018} requires significant manual effort for building a dictionary and designing features. The end-to-end neural network architectures have not achieved competitive results for this task. It is because the data sparseness problem of some NE categories when the size of FG-NER dataset are comparable with NER dataset while the number of NE categories is much larger. Moreover, identifying NE for FG-NER is more difficult compared to NER because these NE types are more complex and longer.
\paragraph{}
Recently, multi-task learning approaches have been proposed for improving the performances of NER systems~\cite{Yang:2017,Lin:2018,Lin:2018b,Changpinyo:2018}. It can be seen as a form of inductive transfer that introduces an auxiliary task as an inductive bias to help a model prefer some hypotheses over the others. Another way to improve NER systems is using contextualized word representations to learn the dependencies among words in a sentence~\cite{Peters:2018}. From these motivations, we investigate the effectiveness of multi-task learning for the end-to-end neural network architecture in both cases uncontextualized and contextualized word representations for FG-NER task.
\paragraph{}
We have novel contributions in two folds. First, to the best of our knowledge, our work is the first study that concentrates on multi-task learning approach for sequence labeling problem in general and for FG-NER task in particular at different aspects including different parameter sharing schemes for multi-task sequence labeling, learning with neural language model, and learning at different word representation settings. We also give empirical analysis to understand the effectiveness of contextualized word representations for FG-NER task. Second, we propose an end-to-end neural network architecture which achieves SOTA result compared to the previous systems that require significant manual effort for building a dictionary and designing features. This neural network system, despite focusing on FG-NER task, can still be applied to any other sequence labeling problems.
\paragraph{}
The remainder of this paper is structured as follows. Section~\ref{sec:approach} describes multi-task learning architectures and contextualized word representations used in our system. Section~\ref{sec:experiments} gives experimental results and discussions. Finally, Section~\ref{sec:conclusion} concludes the paper.
\section{Approach}
\label{sec:approach}
\subsection{Single-Task Sequence Labeling Model}
\paragraph{}
With a recent resurgence of the deep learning approaches, there have been several neural network models proposed for sequence labeling problem. Most of these models shared the same abstract architecture. In particular, each input sentence is fed to these models as a sequence of words and is transformed into a sequence of distributed representations by the word embedding layer. These distributed representations can be improved by incorporating character-level information from Convolutional Neural Network (CNN) or Long Short-Term Memory (LSTM) layer into word embedding layer.
That distributed representation sequence is then passed to the recurrent neural network layer (LSTM or a Gated Recurrent Unit (GRU)), and then a Conditional Random Field (CRF) layer takes as input the output of the recurrent neural network layer to predict the best output sequence~\cite{Huang:2015,Lample:2016,Ma:2016}. 
\paragraph{}
In our work, we re-implement the neural network architecture in~\cite{Ma:2016} which is the combination of CNN, bi-directional LSTM (BLSTM), and CRF models as our base model. 
For training, we minimize the negative log-likelihood function:

\begin{equation}
E = -\sum_{t=1}^{T} \log(P(y_{t}|h_{t}))
\end{equation}
where $h_{t}$ is the output of BLSTM and $y_{t}$ is the label at time step $t$. Decoding can be solved effectively by Viterbi algorithm to find the sequence with the highest conditional probability.

\begin{figure*}[t]
\centering
\begin{subfigure}{0.3\textwidth}
\includegraphics[width=1.1\linewidth]{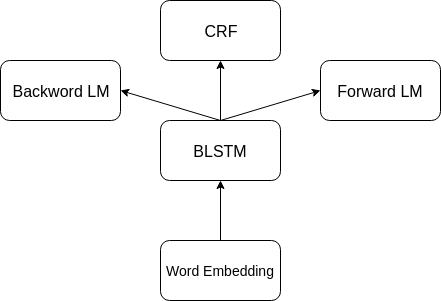}
\subcaption{Single Model (+LM)}
\label{fig:1a}
\end{subfigure}\hspace{0.5\textwidth}
\begin{subfigure}{0.36\textwidth}
\includegraphics[width=1.1\linewidth]{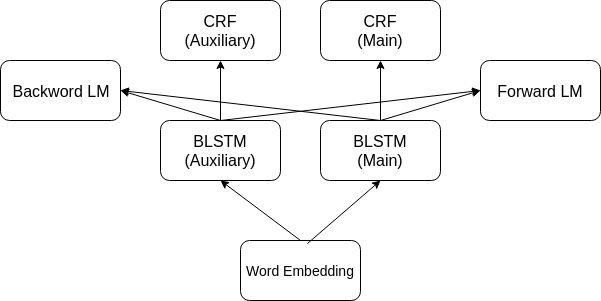}
\subcaption{Embedding-Shared Model (+Shared LM)}
\label{fig:1b}
\end{subfigure}\hspace{0.1\textwidth}
\begin{subfigure}{0.36\textwidth}
\includegraphics[width=1.1\linewidth]{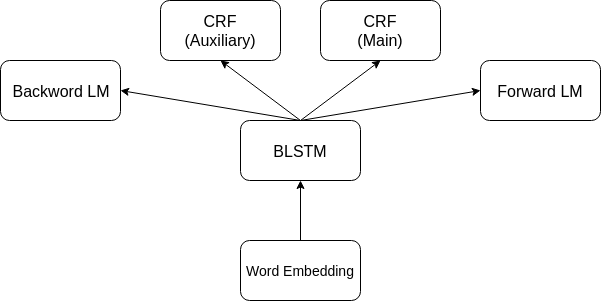}
\subcaption{RNN-Shared Model (+Shared LM)}
\label{fig:1c}
\end{subfigure}\vspace{5mm}
\begin{subfigure}{0.34\textwidth}
\includegraphics[width=1\linewidth]{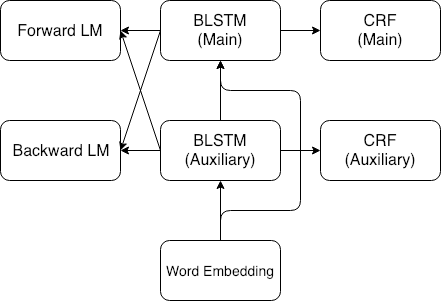}
\subcaption{Hierarchical-Shared Model (+Shared LM)}
\label{fig:1d}
\end{subfigure}\hspace{0.1\textwidth}
\begin{subfigure}{0.34\textwidth}
\includegraphics[width=0.83\linewidth]{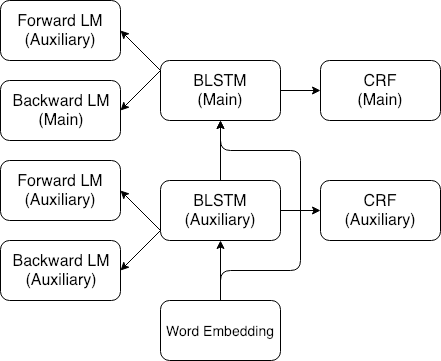}
\subcaption{Hierarchical-Shared Model (+Unshared LM)}
\label{fig:1e}
\end{subfigure}
\caption{Single-Task and Multi-Task Sequence Labeling Models (+LM).}
\label{fig:1}
\end{figure*}
\subsection{Multi-Task Learning with Sequence Labeling Model}
\paragraph{}
Recently, multi-task sequence learning approach has been used successfully in sequence labeling problem~\cite{Yang:2017,Ruder:2017,Peng:2017,Hashimoto:2017,Changpinyo:2018,Clark:2018}. In particular, the main sequence labeling task is learned with auxiliary sequence labeling tasks during training to improve the performance of the main task. These multi-task sequence labeling models are extensions to the base model discussed above with different parameter sharing schemes. 
\paragraph{}
In our work, we investigate two kinds of multi-task sequence labeling models for FG-NER task including same-level-shared model and hierarchical-shared model. To train multi-task sequence labeling models, we minimize both of auxiliary and main objective function. In particular, for an input sequence $\textbf{x} = (x_{1},x_{2},...,x_{T})$, we minimize:
\begin{equation}
E_{auxiliary}=-\sum_{t=1}^{T} \log(P(y^{auxiliary}_{t}|h^{auxiliary}_{t}))
\end{equation}
 if $\textbf{x}$ belongs to auxiliary data, or 
\begin{equation}
E_{main}=-\sum_{t=1}^{T} \log(P(y^{main}_{t}|h^{main}_{t}))
\end{equation}
if $\textbf{x}$ belongs to main data.
$h^{auxiliary}_{t}, h^{main}_{t}$ are the outputs of BLSTM and $y^{auxiliary}_{t}, y^{main}_{t}$ are the labels at time step $t$.
\paragraph{Same-level-Shared Model}
For same-level-shared model, both main and auxiliary tasks are trained and predicted at the same-level layer. Specificially, we experiment with two kinds of same-level-shared model including embedding-shared model (Figure~\ref{fig:1}\subref{fig:1b}) which uses the same embedding layer for both main and auxiliary tasks, and separate LSTM and CRF layers for each task, and RNN-shared model (Figure~\ref{fig:1}\subref{fig:1c}) which uses the same embedding and LSTM layers for both main and auxiliary task, and separate CRF layers for each task.  In RNN-shared model, $\textbf{h}^{auxiliary}, \textbf{h}^{main}$ are the same and are computed from one BLSTM layer:
\begin{equation}
\textbf{h}^{auxiliary} = \textbf{h}^{main} = \textbf{BLSTM}(\textbf{x})
\end{equation}
while in embedding-shared model, $\textbf{h}^{auxiliary}, \textbf{h}^{main}$ are computed from separate BLSTM layers:
\begin{equation}
\textbf{h}^{auxiliary} = \textbf{BLSTM}^{auxiliary}(\textbf{x})
\end{equation}
\begin{equation}
\textbf{h}^{main} = \textbf{BLSTM}^{main}(\textbf{x})
\end{equation}

\paragraph{Hierarchical-Shared Model}
For hierarchical-shared model, we train and predict different supervised tasks at different-level layers. The auxiliary and main tasks are predict by the low-level and high-level layers respectively. To avoid catastrophic interference between main and auxiliary tasks, the word representations are fed into both both low-level and high-level layers. In particular, $\textbf{h}^{auxiliary}, \textbf{h}^{main}$ are computed as follows:
\begin{equation}
\textbf{h}^{auxiliary} = \textbf{BLSTM}^{auxiliary}(\textbf{x})
\end{equation}
\begin{equation}
\textbf{h}^{main} = \textbf{BLSTM}^{main}([\textbf{x};\textbf{h}^{auxiliary}])
\end{equation}

\subsection{Multi-Task Learning with Neural Language Model}
\paragraph{}
Learning with auxiliary sequence labeling task requires additional data which may not be available for some languages. For this reason, several models have been proposed for training sequence labeling task with other unsupervised learning tasks. In particular,~\cite{Cheng:2015,Rei:2017} trained single-task sequence labeling models with neural language model simultaneously.
\paragraph{}
In our work, we incorporate a word-level neural language model into both single and multi-task sequence labeling models to improve the performances. Specifically, we put the hidden state from BLSTM at each time step into softmax layer to predict the next and previous words. Note that we use two separate language models for each forward and backward passes of BLSTM. The objective function now is the combination of sequence labeling and language model objective functions and is computed as follows:

\begin{equation}
E_{joint} = E +  \lambda(\overleftarrow{E}_{LM} + \overrightarrow{E}_{LM})
\end{equation}
where $\lambda$ is a parameter controlled the impact of the language modeling task to the sequence labeling task and $\overleftarrow{E}_{LM}, \overrightarrow{E}_{LM}$ are the objective functions of forward and backward language models. These objective functions are computed as follows:

\begin{equation}
\overleftarrow{E}_{LM} = -\sum_{t=1}^{T} \log(P(w_{t-1}| \overleftarrow{h_{t}}))
\end{equation}
\begin{equation}
\overrightarrow{E}_{LM} = -\sum_{t=1}^{T} \log(P(w_{t+1}| \overrightarrow{h_{t}}))
\end{equation}
where $\overrightarrow{h_{t}}, \overleftarrow{h_{t}}$ are hidden states of forward and backward LSTM and $w_{0}, w_{T+1}$ are special tokens \textit{START, END}. We investigate two kinds of incorporating neural language model into our multi-task sequence labeling model: shared-LM which  shares neural language model for both auxiliary and main sequence labeling task and unshared-LM which uses separate neural language model for each task. Figure~\ref{fig:1}\subref{fig:1d} and Figure~\ref{fig:1}\subref{fig:1e} show the difference between these two kinds of incorporating neural language model.

\subsection{Deep Contextualized Word Representations}
\paragraph{}
Uncontextualized word embeddings such as Word2Vec~\cite{Mikolov:2013}, GloVe~\cite{Pennington:2014} have been used widely in neural natural language processing models and have improved their performances. However, these word embeddings still have some drawbacks. In particular, it is difficult for them to represent the complex characteristics of a word and its meaning at different contexts. Recently, deep contextualized word representations have been proposed to solve these problems.~\cite{Peters:2018} introduces the word representations which are computed from multi-layer bidirectional language model with character convolutions. Unlike ~\cite{Peters:2018},~\cite{Radford:2018} use Transformer instead of BLSTM to calculate the language model.~\cite{Devlin:2018} improves \cite{Radford:2018} work by jointly learning both left and right context in Transformer.
\paragraph{}
Our proposed multi-task models can be trained with any kind of these contextualized word representations but in the scope of this paper, we only experiment with contextualized word representations described in~\cite{Peters:2018} which are called Embeddings from Language Models (ELMo) and leave other contextualized word representations in our future works. ELMo are functions of the entire input sentence and are computed as follows:

\begin{equation}
\textbf{ELMo}_{t} = \gamma \sum_{l=1}^{L}s_{l}h_{t,l}^{LM}
\end{equation}
where $h_{t,0}^{LM}$ is the input representation and $h_{t,l}^{LM}$ is the output at $l^{th}$ layer of L-layer bidirectional language model at time step $t$, $\textbf{s} = (s_{1},s_{2},...,s_{L})$ are the softmax-normalized weights and $\gamma$ are the scalar parameter which allows the model to scale the ELMo vector. In our work, we incorporate a 2-layer bidirectional language model pre-trained on 1 Billion Word Language Model Benchmark dataset to our system. We set $s_{1} = 0$ and $s_{2} = 1$ which means we only use the output of $2^{nd}$ layer of the bidirectional language model as an input for the next layer in our system.

\section{Experiments}
\label{sec:experiments}
\subsection{Datasets}
\paragraph{}
We conduct our experiments with FG-NER as our main task and POS tagging, chunking, NER, and language model as our auxiliary tasks. For FG-NER task, we use the English part of the dataset described in~\cite{Nguyen:2017,Mai:2018}. For chunking task, we use CoNLL 2000 dataset~\cite{Tjong:2000}. This dataset has only training and testing sets so we used one part of the training set for validation. For NER task, we use CoNLL 2003 and OntoNotes 5.0 datasets~\cite{Tjong:2003,Pradhan:2012}. OntoNotes 5.0 dataset is also used for POS tagging task. The details of each dataset are described in Table~\ref{tab:1}.

\begin{table}[]
\centering
\resizebox{0.45\textwidth}{!}{
\begin{tabular}{lrrrrrr}
\toprule
\multirow{2}{*}{Datasets} & \multicolumn{3}{c}{\#Sentence} & \multirow{2}{*}{\#Word} & \multirow{2}{*}{\#Label} \\ 
 & Train & Dev & Test &  &  \\ \midrule
FG-NER & 14176 & 1573 & 3942 & 32052 & 208 \\ 
POS & 58891 & 8254 & 6457 & 68241 & 51 \\ 
Chunk & 8000 & 936 & 2012 & 21589 & 23 \\ 
NER (CoNLL) & 14987 & 3466 & 3684 & 30290 & 8 \\ 
NER (OntoNotes) & 58891 & 8254 & 6457 & 68241 & 30 \\ \bottomrule
\end{tabular}}
\caption{Statistics of datasets used in our experiments.}
\label{tab:1}
\end{table}

\begin{table}[t]
\centering
\resizebox{0.35\textwidth}{!}{\begin{tabular}{llr}
\toprule
                                                 & Hyper-parameter        & Value \\ \midrule
\multicolumn{1}{l}{LSTM}      & hidden size            & 256   \\  
 \midrule
\multicolumn{1}{l}{\multirow{2}{*}{CNN}}       & window size            & 3     \\ 
\multicolumn{1}{l}{}                           & \#filter               & 30    \\ \midrule
\multicolumn{1}{l}{\multirow{2}{*}{Dropout}}   & input dropout          & 0.33  \\ 
\multicolumn{1}{l}{}                           & BLSTM dropout          & 0.5   \\ \midrule
\multicolumn{1}{l}{\multirow{3}{*}{Embedding}} & GloVe dimension        & 300   \\ 
\multicolumn{1}{l}{}                           & ELMo dimension         & 1024  \\ 
\multicolumn{1}{l}{}                           & $\gamma$         & 1  \\ \midrule
\multicolumn{1}{l}{Language Model}             & $\lambda$ & 0.05  \\ \midrule
\multicolumn{1}{l}{\multirow{3}{*}{Training}}  & batch size             & 16    \\  
\multicolumn{1}{l}{}                           & initial learning rate  & 0.01  \\ 
\multicolumn{1}{l}{}                           & decay rate             & 0.05  \\ \bottomrule
\end{tabular}}
\caption{Hyper-parameters used in our systems.}
\label{tab:2}
\end{table}

\subsection{Training and Evaluation Method}
\paragraph{}
The training procedure for multi-task sequence labeling models is as follows. For same-level-shared models, at each iteration, we first sample a task (main or auxiliary tasks) by Bernoulli trial based on sizes of datasets. Next, we sample a batch of training examples from the given task and then update gradients for both the shared parameters and the task-specific parameters according to the loss function of the given task. For hierarchical-shared models, at each iteration, we train the auxiliary (low-level) task first and then move to the main (high-level) task because selecting the task randomly hampers the effectiveness of hierarchical-shared models~\cite{Hashimoto:2017}. We use stochastic gradient descent algorithm with decay rate $0.05$. Table~\ref{tab:2} shows the hyper-parameters we used in our models. 
\paragraph{}
We evaluate the performance of our system with $F_{1}$ score:
\begin{equation*}
F_{1} = \frac{2*\mathtt{precision}*\mathtt{recall}}{\mathtt{precision}+\mathtt{recall}}
\end{equation*}
Precision and recall are the percentage of correct named entities identified by the system and the percentage of identified named entities present in the corpus respectively. To compare fairly with previous systems, we use an available evaluation script provided by the CoNLL 2003 shared task\footnote{\url{http://www.cnts.ua.ac.be/conll2003/ner/}} to calculate $F_{1}$ score of our FG-NER system.
\subsection{Results}
\begin{table*}[!t]
\centering
\resizebox{0.85\textwidth}{!}{\begin{tabular}{lrrrrr}
\toprule
Model & FG-NER & +Chunk & \begin{tabular}[c]{@{}c@{}}+NER \\ (CoNLL)\end{tabular} & +POS & \begin{tabular}[c]{@{}c@{}}+NER \\ (Ontonotes)\end{tabular} \\ \midrule
Base Model (GloVe)& 81.51 & - & - & - & - \\ 
RNN-Shared Model (GloVe) & - & 80.53 & 81.38 & 80.55 & 81.13 \\ 
Embedding-Shared Model (GloVe) & - & 81.49 & 81.21 & 81.59 & 81.24 \\  
Hierarchical-Shared Model (GloVe) & - & 81.65 & \textbf{82.14} & 81.27 & 81.67 \\ \midrule
Base Model (ELMo) & 82.74 & - & - & - & - \\ 
RNN-Shared Model (ELMo) & - & 82.60 & 82.09 & 81.77 & 82.12 \\ 
Embedding-Shared Model (ELMo) & - & 82.75 & 82.45 & 82.34 & 81.94 \\
Hierarchical-Shared Model (ELMo) & - & \textbf{83.04} & 82.72 & 82.76 & 82.96 \\  \midrule
Base Model (GloVe) + LM~\cite{Rei:2017} & 81.77 & - & - & - & - \\ 
RNN-Shared Model (GloVe) + Shared-LM & - & 80.83 & 81.34 & 80.69 & 81.45 \\ 
Embedding-Shared Model (GloVe) + Shared-LM & - & 81.54 & 81.95 & 81.86 & 81.34 \\ 
Hierarchical-Shared Model (GloVe) + Shared-LM & - & 81.69 & \textbf{81.96} & 81.42 & 81.78 \\ \midrule
Base Model (ELMo) + LM & 82.91 & - & - & - & - \\ 
RNN-Shared Model (ELMo) + Shared-LM & - & 82.68 & 82.64 & 81.61 & 82.36 \\ 
Embedding-Shared Model (ELMo) + Shared-LM & - & 82.61 & 82.32 & 82.46 & 82.45 \\ 
Hierarchical-Shared Model (ELMo) + Shared-LM & - & 82.87 & 82.82 & 82.85 & \textbf{82.99} \\ \midrule
Hierarchical-Shared Model (GloVe) + Unshared-LM & - & 81.77 & 81.80 & 81.72 & 81.88 \\
Hierarchical-Shared Model (ELMo) + Unshared-LM & - & \textbf{83.35} & 83.14 & 83.06 & 82.82 \\ \midrule
\cite{Mai:2018} & 83.14 & - & - & - & - \\ \bottomrule
\end{tabular}}
\caption{Results in $F_{1}$ scores for FG-NER (We run each setting five times and report the average $F_{1}$ scores.)}
\label{tab:3}
\end{table*}
\paragraph{Base Model}
Our base model is similar to LSTM + CNN + CRF model in~\cite{Mai:2018}, but in contrast to their model, we implement by PyTorch instead of Theano and train sentences with same length at each batch to make the training process faster. It achieves $F_{1}$ score of $81.51\%$ compared to $80.93\%$ reported in their paper.
\paragraph{Deep Contextualized Word Representations}
In the first experiment, we investigate the effectiveness of contextualized word representations (ELMo) compared to uncontextualized word representations (GloVe) when incorporating in our FG-NER systems (\textit{\textbf{Base Model (GloVe)}} vs. \textit{\textbf{Base Model (ELMo)}}). From Table~\ref{tab:3}, we see that using ELMo significantly improves the $F_{1}$ score of our system compared to using GloVe (from $81.51\%$ to $82.74\%$).
\paragraph{}
To further investigate this phenomenon, we give an analysis to see which NE types are improved when using ELMo. Table~\ref{tab:4} shows $F_{1}$ scores of 5 NE types which are most improved and their average token lengths. While the average token length of NEs in our dataset is $1.9$, the average token lengths of these NE types are much longer. It shows that ELMo helps to improve the performance of our system when identifying NEs which are long sequences. This result is understandable because \textit{\textbf{Base Model (GloVe)}} relies on only BLSTM layer to learn the dependencies among words in sequence to predict NE labels while \textit{\textbf{Base Model (ELMo)}} learns these dependencies by both embedding and BLSTM layers. Unlike NER, NE types in FG-NER are often more complex and longer so using only BLSTM layer is not sufficient to capture these dependencies.

\begin{table}[]
\centering
\resizebox{0.45\textwidth}{!}{\begin{tabular}{lrrr}
\toprule
Named Entity       & GloVe & ELMo  & Token Length \\ \midrule
Book               & 48.65 & 76.92 & 3.2          \\ 
Printing Other     & 60.38 & 83.33 & 3.5          \\
Spaceship          & 61.90 & 80.00 & 2.7          \\ 
Earthquake         & 75.00 & 90.20 & 3.8          \\ 
Public Institution & 80.00 & 95.00 & 4.2          \\ \bottomrule
\end{tabular}}
\caption{5 most improved NE types when using ELMo.}
\label{tab:4}
\end{table}


\paragraph{Parameter Sharing Schemes}
In the second experiment, we investigate the impact of training FG-NER with other auxiliary sequence labeling tasks including POS tagging, chunking, and NER by our multi-task sequence labeling models at different parameter sharing schemes. In particular, we compare three kinds of multi-task sequence labeling architectures including embedding-shared, RNN-shared, and hierarchical-shared models. The size of the original OntoNotes dataset is much larger than FG-NER dataset so it is difficult for our system to focus on learning FG-NER task. Thus, we sample 10,000 sentences from OntoNotes for training POS tagging and NER.
\paragraph{}
Table~\ref{tab:3} shows the performances of our multi-task sequence labeling models with GloVe and ELMo representations. In both cases, hierarchical-shared model gives the best performances. In particular, it achieves an $F_{1}$ score of $82.14\%$ when learning with NER (CoNLL) and an $F_{1}$ score of $83.04\%$ when learning with NER (Ontonotes) compared to $F_{1}$ scores of $81.51\%$ and $82.74\%$ of base model in GloVe and ELMo settings respectively. For same-level-shared models, they also achieve better results compared to base model but the differences are not very large. These results indicate that learning FG-NER with other sequence labeling tasks at different parameter sharing schemes helps to improve the performances of FG-NER system. Also, in most cases, it is more beneficial when learning the auxiliary and the main tasks at different levels (\textit{\textbf{hierarchical-shared model}}) compared to learning at the same level (\textit{\textbf{RNN-shared and embedding-shared models}}).
\paragraph{}
For same-level sharing scheme, we also see that \textit{\textbf{embedding-shared model}} achieves better performances than \textit{\textbf{RNN-shared model}} in most cases. The gap between these two models is larger when the auxiliary task is more different from the main task (POS tagging, chunking are more different from FG-NER compared to NER).

\paragraph{Neural Language Model}
In the third experiment, we incorporate our systems including both single and multi-task sequence labeling models with neural language model. We experiment with two kinds of incorporating neural language model: shared-LM which shares neural language model for both auxiliary and main sequence labeling tasks and unshared-LM which uses separate neural language model for each task. For single-task model, incorporating neural language model helps to improve performance from $81.51\%$ to $81.77\%$ and from $82.74\%$ to $82.91\%$ in GloVe and ELMo settings respectively. For multi-task models, with shared-LM, our best result is an $F_{1}$ score of $82.99\%$ when learning hierarchical-shared FG-NER model with NER (Ontonotes), and with unshared-LM, our best result is an $F_{1}$ score of $83.35\%$ when learning hierarchical-shared FG-NER model with chunking. We also see that using unshared-LM helps our multi-task models achieves better performances compared to using shared-LM in most cases.

\paragraph{Comparison with SOTA System}
Our best system achieves the SOTA result for FG-NER. In particular, our hierarchical-shared model with chunking as an auxiliary sequence labeling task and unshared-LM achieves an $F_{1}$ score of $83.35\%$ compared to $83.14\%$ of the previous SOTA model for FG-NER~\cite{Mai:2018}. While that model requires significant manual effort for building a dictionary and designing hand-crafted features, our best model is truly end-to-end framework without using any additional information.

\section{Conclusion}
\label{sec:conclusion}
We present an experimental study on the effectiveness of using multi-task learning with contextualized word representations in FG-NER task. In particular, we examine the multi-task approach at different aspects including different parameter sharing schemes for multi-task sequence labeling, learning with neural language model, and learning at different word representation settings. Our best model, while does not use any additional manual effort for creating data and designing features, achieves an $F_{1}$ score of $83.35\%$ which is the SOTA result compared to the previous FG-NER model.

\bibliography{ijcai19}
\bibliographystyle{named}

\end{document}